%% file: neurips_2023.tex
\documentclass{article}

\PassOptionsToPackage{numbers}{natbib}

\usepackage[preprint]{neurips_2023}

\usepackage[utf8]{inputenc} 
\usepackage[T1]{fontenc}    
\usepackage[hidelinks]{hyperref}       
\usepackage{url}            
\usepackage{booktabs}       
\usepackage{amsfonts}       
\usepackage{nicefrac}       
\usepackage{microtype}      
\usepackage{xcolor}         
\usepackage{amsmath}
\usepackage{amssymb}
\usepackage{float}
\usepackage{graphicx}
\usepackage{subfig}
\usepackage{xspace}
\usepackage{bbm}
\usepackage{tabularx}
\usepackage{comment}
\usepackage[linesnumbered,ruled,vlined]{algorithm2e}

\SetCommentSty{mycommfont}
\SetKwInput{KwInput}{Input}                
\SetKwInput{KwOutput}{Output}              
\newcommand{\etal}{\textit{et al}.}

\newcommand{\eg}{\textit{e}.\textit{g}.}
\DeclareMathOperator*{\argmax}{arg\,max}
\DeclareMathOperator*{\argmin}{arg\,min}
\def\shownotes{1}  \ifnum\shownotes=1
\newcommand{\authnote}[2]{$\ll$\textsf{\footnotesize #1 notes: #2}$\gg$}
\else
 \newcommand{\authnote}[2]{}
\fi

\DeclareUnicodeCharacter{2212}{-}

\newcommand{\seqseq}{\textsc{Seq2Seq}\xspace}
\newcommand{\attention}{\textsc{Attention}\xspace}
\newcommand{\transformer}{\textsc{Transformer}\xspace}

\newcommand{\rnndecompilation}{\textsc{RNNDecompilation}\xspace}
\newcommand{\neuraldecompilation}{\textsc{NeuralDecompilation}\xspace}
\newcommand{\treetotree}{\textsc{Tree2Tree}\xspace}
\newcommand{\slm}{\textsc{SLM}\xspace}
\newcommand{\gmg}{\textsc{GMG}\xspace}
\newcommand{\lpn}{\textsc{LPN}\xspace}
\newcommand{\snm}{\textsc{SNM}\xspace}
\newcommand{\lrpg}{\textsc{LRPG}\xspace}
\newcommand{\lcpc}{\textsc{LCPC}\xspace}
\newcommand{\ptrnet}{\textsc{PtrNet}\xspace}
\newcommand{\ptrgen}{\textsc{PtrGenNet}\xspace}
\newcommand{\copynet}{\textsc{CopyNet}\xspace}
\newcommand{\coda}{\textsc{Coda}\xspace}
\newcommand{\neutron}{\textsc{Neutron}\xspace}
\newcommand{\codex}{\textsc{Codex}\xspace}
\newcommand{\alphacode}{\textsc{AlphaCode}\xspace}
\newcommand{\wordtovec}{\textsc{Word2Vec}\xspace}
\newcommand{\codetovec}{\textsc{Code2Vec}\xspace}
\newcommand{\codetoseq}{\textsc{Code2Seq}\xspace}
\newcommand{\coderl}{\textsc{CodeRL}\xspace}
\newcommand{\codetfive}{\textsc{CodeT5}\xspace}

\newcommand{\llm}{\textsc{LLM}\xspace}

\newcommand{\conversation}{\textsc{Conversation}\xspace}
\newcommand{\repl}{\textsc{REPL}\xspace}

\newcommand{\gpt}{\textsc{GPT}\xspace}
\newcommand{\cubert}{\textsc{CuBERT}\xspace}
\newcommand{\codebert}{\textsc{CodeBERT}\xspace}
\newcommand{\intellicode}{\textsc{IntelliCode}\xspace}
\newcommand{\plbart}{\textsc{PLBART}\xspace}
\newcommand{\graphcodebert}{\textsc{GraphCodeBERT}\xspace}
\newcommand{\langagnostic}{\textsc{LangAgnostic}\xspace}
\newcommand{\dobf}{\textsc{DOBF}\xspace}
\newcommand{\deepfix}{\textsc{DeepFix}\xspace}
\newcommand{\synfix}{\textsc{SynFix}\xspace}
\newcommand{\semfix}{\textsc{SemFix}\xspace}
\newcommand{\incoder}{\textsc{InCoder}\xspace}
\newcommand{\polycoder}{\textsc{PolyCoder}\xspace}
\newcommand{\asn}{\textsc{ASNs}\xspace}
\newcommand{\cnndecoder}{\textsc{CNNDecoder}\xspace}
\newcommand{\pixtwocode}{\textsc{pix2Code}\xspace}
\newcommand{\seqtwosql}{\textsc{Seq2SQL}\xspace}
\newcommand{\rlcorrection}{\textsc{RLCorrection}\xspace}
\newcommand{\sparsepointernetwork}{\textsc{SparsePointerNetwork}\xspace}
\newcommand{\nltwobash}{\textsc{NL2Bash}\xspace}
\newcommand{\nltwocode}{\textsc{NL2Code}\xspace}
\newcommand{\compcoder}{\textsc{CompCoder}\xspace}
\newcommand{\pointermixture}{\textsc{PointerMixture}\xspace}
\newcommand{\treegen}{\textsc{TreeGen}\xspace}

\newcommand{\dreamcoder}{\textsc{DreamCoder}\xspace}

\title{Neural Machine Translation for Code Generation}

\author{%
  Dharma KC \\
  University of Arizona\\
  \texttt{kcdharma@arizona.edu}
  \And
  Clayton T. Morrison \\
  University of Arizona \\
  \texttt{claytonm@arizona.edu}
}

\begin{document}

\maketitle

\input{abstract}
\input{introduction}
\input{overview}
\input{nmt}

\section{Neural Machine Translation for Code: NMT4Code}
\label{section:nmt4code}
In this section, we summarize papers that use neural machine translation for code generation based on the output representation produced by these methods and the methods they use for code generation. These papers can be summarized into two groups namely \emph{Sequence} and \emph{Graph} based on the output representation they produce for the source code. 
%


Sequence methods represent the output source code in a sequential representation. This sequential representation can be either a sequential form of the original source code or the linearized form of the AST representation of the source code. The advantage of sequential representation is that we can use sequence-to-sequence models~\cite{sutskever2014sequence, bahdanau2014neural, vaswani2017attention} in their raw form for the generation of the source code.

\input{sourcecode}

\input{astsequence}

\subsection{Graph}
These methods generate a graph (AST) representation of the source code. The advantage of such methods is that the AST captures the syntax of the given language explicitly. The major disadvantage of such systems is that it's non-trivial to parallelize such systems and we need another extra module that generates source code from AST representation. These methods generate a graph (AST) representation of the source code. 

\input{binarytree}
\input{narytree}
\input{copy}
\input{repr}
\input{open}
\input{dataset}
\input{evaluation}
\input{conclusion}


\bibliographystyle{plainnat}
\bibliography{main}

\end{document}

%% file: abstract.tex
\begin{abstract}
    Neural machine translation (NMT) methods developed for natural language processing have been shown to be highly successful in automating translation from one natural language to another. Recently, these NMT methods have been adapted to the generation of program code. In \emph{NMT for code generation}, the task is to generate output source code that satisfies constraints expressed in the input. In the literature, a variety of different input scenarios have been explored, including generating code based on natural language description, lower-level representations such as binary or assembly (neural decompilation), partial representations of source code (code completion and repair), and source code in another language (code translation). In this paper we survey the NMT for code generation literature, cataloging the variety of methods that have been explored according to input and output representations, model architectures, optimization techniques used, data sets, and evaluation methods. We discuss the limitations of existing methods and future research directions.
\end{abstract}

%% file: introduction.tex
\section{Introduction}
Neural machine translation (NMT) has been widely used in natural language processing, where the task is to translate sentences in one language to sentences in another language (such as from English to German). Most NMT frameworks follow an Encoder-Decoder architecture pioneered by Sutskever \etal~\cite{sutskever2014sequence}. In this architecture, input sequences of tokens (representing words or characters in the source language) are input to an encoder neural network that computes an internal representation, and then a decoder neural network takes that internal representation and decodes it into an output sequence of tokens corresponding to the target language. Recent work has begun to explore applying this and related ideas to the task of generating source code, leading to a new sub-field of NMT for code generation. This is now an active and rapidly evolving area of research with many applications~\cite{allamanis2018survey, le2020deep, gulwani2017program}. For example, code generation from natural language can help novice developers to write code and seasoned programmers to work more efficiently ~\cite{li2022competition, austin2021program}. In so-called neural decompilation, NMT for code generation has been applied to translate from assembly or binary to source code. This in turn can help security researchers with malware identification, understanding the functionality of programs, program comparison, and more~\cite{fu2019coda, liang2021neutron}. Code generation from partial source code can help in program completion, repair, and providing automated feedback~\cite{bhatia2016automated, devlin2017semantic}. And finally, in the setting directly analogous to NMT for natural language translation, NMT methods can be used to translate from source code in one programming language to another~\cite{chen2018tree}. In the literature, researchers are using various methods in each of these application domains, but in general, the field is evolving very fast, with important results being siloed within particular application communities, and important insights that might transfer to other application is challenging to track. Summarizing these papers based on the methods and ideas they use and identifying their limitations is important to inform the development of new architectures and solutions. In the survey presented here, we develop several dimensions along which to distinguish the various approaches. One natural dimension is to consider the class of neural network architecture that is used; this includes variants of recurrent neural networks (RNNs), transformers and large language models (LLMs), tree decoders, graph neural networks (GNNs), neurosymbolic methods, and incorporation of reinforcement learning (RL). We consider each in terms of their advantages and limitations. Along another dimension, we characterize the approaches according to the
output representation they generate: sequence or graph (abstract syntax tree (AST)) with advantages and limitations of each of the methods. We hope the contents of this paper will be useful for new researchers to get an overview of the field, and for experts to design new architectures and solutions suitable for their tasks and requirements. When the boundary between methods is not clear or when papers use hybrid methods, we have grouped them based on their major component. \\

Program Synthesis: foundations and trends~\cite{gulwani2017program} provides a detailed analysis of traditional methods such as inductive program synthesis (enumerative search or sketch-based solutions~\cite{solar2009sketching}), and deductive program synthesis techniques. Deductive program synthesis works by translating the descriptions from formal language (specifications) to the source code using theorem prover to find a proof that satisfies all the constraints and extracts the program from the proofs~\cite{green1981application, manna1971toward}. Thus, it requires explicit program descriptions in the formal language to start with. Inductive methods~\cite{gulwani2011automating} can generate source code from input-output example pairs, but they perform a search over all the language rules and are hard to scale to real-world source code generation. They mostly work with domain-specific languages (DSLs) with few rules compared to the grammar of general programming languages such as C, C++, and Python. Existing survey papers~\cite{allamanis2018survey, le2020deep} also provide a survey of probabilistic methods, domain-specific language (DSL) guided models, and n-gram language models~\cite{hindle2016naturalness} along with applications of machine learning techniques on code such as code summarization, bug fixing, and more. Recent paper~\cite{le2020deep} covers deep learning-based methods for source code generation but does not cover the latest techniques like RL-based methods, papers from the neural decompilation domain, recent large language models like \codex and \alphacode, and code representation techniques. Another line of work popular for program synthesis is based on differentiable interpreters. Differential interpreter based methods~\cite{bovsnjak2017programming, ling2016latent, graves2014neural, kurach2015neural, kaiser2015neural, reed2015neural, graves2016hybrid} generate source code from input output examples. They define a differentiable mapping from inputs and source code to the outputs and use gradient descent to search for the best program that satisfies the constraints. The major disadvantage of differentiable interpreter-based methods is that each problem is solved independently, and existing methods based on this idea do not yet scale to code generation in a general programming language such as C, C++, and Python. We do not cover methods that do not yet scale to code generation in general programming languages such as C, C++, and Python and suggest our readers look into survey papers. Semantic parsing and code generation are quite related tasks where the ideas from one field can be transferred to another, we refer our readers to this survey paper~\cite{lee2021toward} on semantic parsing and code generation. Similarly, ~\cite{wu2022survey} is more focused on code understanding while we focus on code generation. \\

This survey paper is organized as follows: section overview[\ref{section:overview}] provides brief overview of the components used by NMT based algorithms, section NMT [\ref{section:nmt}] introduces neural machine translation framework followed by methods used in NLP for language translation that serve as a basis for current code generation techniques, section NMT4Code [\ref{section:nmt4code}] provides description of methods used by existing code generation papers, section copy mechanism [ref{section:copy}] introduces copy mechanism that will help the NMT models to copy some tokens from the input sequence to the output sequence, section representation learning [\ref{section:representation}] provides current representation learning techniques, section datasets [\ref{section:datasets}] provides a list of datasets being used for source code generation, section evaluation [\ref{section:evaluation}] provides evaluation techniques, section open problems [\ref{section:open}] provides open problems and future research directions. Finally, in section conclusion [\ref{section:conclusion}] we conclude our paper.

%% file: overview.tex
\section{Overview}
\label{section:overview}
In this section, we provide a summary of the important components used by these code generation papers.

\subsection{Recurrent neural networks (RNNs)}
Recurrent neural networks~\cite{rumelhart1985learning, jordan1997serial, siegelmann1993foundations} are the extension of feedforward models for learning from a sequence. At any time step $t$, given the current input token ($x_t$) and previous hidden state ($x_{t-1}$), the RNN unit produces a hidden state ($h_{t}$: summary of the sequence up to current time step) along with some output ($o_t$).
 We can model $o_t$ and $h_t$ with following equations:
\begin{align}
    h_t &= tanh( W_i * x_t + W_h * h_{t-1} ) \\
    o_t &= W_o * h_t
\end{align}
Given a sequence of words ($x_1, x_2, ..., x_t)$, RNN works in a sequential fashion producing output ($o_t$) and hidden state ($h_t$) at each time step $t$. The advantage of RNN is that it can handle a sequence of any length. The disadvantage of RNN is that it can not handle long-range dependencies, and is hard to train because of the vanishing gradient problem~\cite{pascanu2013difficulty, hochreiter1998vanishing}. Long short-term memory (LSTM)~\cite{hochreiter1997long} and gated recurrent units (GRUs)~\cite{cho2014properties} are the variations of RNNs with a gated mechanism with two states: hidden state for short-term memory and cell state for long term memory, and are shown to handle long-range dependencies better than vanilla RNNs. We can use RNNs and their variants to generate the representation for sequential data like a sequence of source code, a linearized form of the abstract syntax tree (AST), or a natural language description of the source code.

\subsection{Transformer and large language models (LLMs)}
Transformer~\cite{vaswani2017attention} removes the sequential encoding mechanism of RNNs and instead encodes tokens in feedforward networks with multi-headed dot product attention layers. The advantage of transformers is that they can handle long-range dependencies better than RNNs and are easy to parallelize.
Given, query ($Q$), value ($V$), and key ($K$) vectors, The attention is then calculated as follows:

\begin{align}
    MultiHeadAttention (Q, K, V) &= concat([head_1, head_2, ...., head_k] W_0 \\
    head_i &= Attention(QW_i^{Q}, KW_i^{K}, VW_i^{V})\\
    Attention(Q, K, V) &= softmax(\frac{QK^{T}}{\sqrt{d_k}})V
\end{align}

Where $\sqrt{d_k}$ is the scaling factor equal to the feature vector dimension. The input is transformed into a key, query, and value vector using a linear operator and the attention is calculated using multiple heads as shown by the above equations. The feature vector for a query token is updated with the linear combination of the feature vector of other tokens. This feature vector then goes through another feedforward layer with a residual connection and this process is repeated. Given that the source code has long-range dependencies~\cite{le2020deep} and transformers can handle long-range dependencies, transformers are important architecture for source code generation. Large language models based on Transformer-like GPT-3~\cite{brown2020language} and BERT~\cite{devlin2018bert} have dominated the field of NLP for sequence generation and representation learning respectively, and are being used increasingly for code generation~\cite{wang2021codet5, feng2020codebert}. We can use transformers or LLMs to generate the sequential representation of the source code, or the linearized form of the abstract syntax tree (AST).

\subsection{TreeDecoder}
Tree decoding-based methods such as Tree-to-Tree translation~\cite{chen2018tree} generate a tree by decoding a node at a time with some tree traversal algorithms such as depth-first search (DFS) or breadth-first search (BFS). Tree-based decoders generally use attention~\cite{bahdanau2014neural} over the input sequence to generate a new node. Tree-based decoders also use parent feeding (using the hidden state of the parent)~\cite{chen2018tree} to improve the decoding of a new node. The system can be extended to attend to the partially generated tree along with attention to the input sequence. The core components of these decoders are RNNs (LSTMs, and GRUs). For example, Tree-to-Tree~\cite{chen2018tree} uses two LSTMs, one for the left child and another for the right child prediction. Most of the tree decoder-based code generation methods first generate an abstract syntax tree (AST) that can then be converted into the source code.

\subsection{GNNs}
Graph neural networks~\cite{wu2020comprehensive, kipf2016semi, velivckovic2017graph, dharma2023texture} are quite powerful methods for graph-structured prediction and learning. Most of them work on the principle of message passing where a node gets some information about its neighbors and updates its state. Let's consider our partial AST as a graph $\mathcal{G} = (\mathcal{V}, \mathcal{E})$, where $\mathcal{V}$ is the set of nodes and $\mathcal{E}$ is the set of edges. Let, $\mathcal{X} \in \mathbb{R}^{|v| * d}$ be the set of node features where each node $v \in \mathcal{V}$ has a $d$ dimensional feature. The $k^{th}$ message passing iteration of a GNN can be modeled as a variation of the following equation~\cite{grlbook}:

\begin{align}
    h_v^{(k+1)} &= \texttt{update}^{(k)}(h_v^{(k)}, \texttt{aggregate}^{(k)}(h_u^{(k)}, \quad \forall u \in \mathcal{N}(v))\\
    &= \texttt{update}^{(k)}(h_v^{(k)}, \boldsymbol{m}_{\mathcal{N}(v)}^{(k)})
\end{align}

Where $\mathcal{N}(v)$ denotes the neighbors of node $v$. At any iteration of the GNN, the \texttt{aggregate} function takes the embedding of the neighbors of the node $v$ and combines them into one embedding vector. The \texttt{update} function takes the embedding of the node $v$ at the previous time-step and the output embedding vector of the  \texttt{aggregate} function to give us a new embedding for the node $v$. Here, \texttt{update} and \texttt{aggregate} can be any differentiable functions. Given that the \texttt{aggregate} function takes a set of inputs, GNNs defined this way are permutation invariant. Moreover, the equations can be easily modified to include edge embedding vector information if available. Note that the one iteration of the GNN collects information from the first-hop neighborhood. If we want to collect information from the k hop neighborhood, we can run the GNN k times [k layers]. We can use GNNs to generate the tree representation of the source code namely abstract syntax tree (AST).

\subsection{Neurosymbolic}
Neurosymbolic methods~\cite{garcez_neurosymbolic_2023, chaudhuri2021neurosymbolic} are methods that combine neural networks with symbolic logic. In this survey paper, we include papers that have a neural component and a symbolic component such as DeepCoder~\cite{balog2016deepcoder} and DreamCoder~\cite{ellis2020dreamcoder} as neurosymbolic methods. Another paradigm of neurosymbolic methods generates a program $P$ that when applied to the input produces the desired output. For example, Fix Bugs with Transformer through a Neural-Symbolic Edit Grammar~\cite{hu2022fix} uses transformers~\cite{vaswani2017attention} and pointer networks~\cite{vinyals2015pointer} to predict a sequence of edit grammar rules (DSL rules such as insert and delete from a location). Let's call these sequences of actions a program ($P$). This program ($P$) when applied to the buggy source code ($x$), outputs the bug-free program ($y$). The advantage of neurosymbolic methods is that the DSL rules are humanly interpretable, and also the method can generalize better at extrapolation than compared to neural-only approaches. The neurosymbolic methods are divided into multiple groups based on their major component whether it is a neural or symbolic module~\cite{garcez_neurosymbolic_2023}. The more symbolic components the method uses, the more interpretable it is but harder to scale to large code generation problems.

\subsection{RL}
All of the above methods do not take advantage of the unit tests and compiler error messages as these are non-differentiable. Algorithms such as REINFORCE~\cite{williams1992simple} can be used to train the system from such non-differentiable values. Reinforcement learning-based methods convert the results of unit tests and compiler output to reward values and mostly use policy gradient methods~\cite{sutton1999policy, dharma2021policy} to train the network based on these signals~\cite{le2022coderl}. Once we have a mechanism to calculate reward based on these signals, we can use it as a loss function where our objective is to maximize the total reward. Thus the loss becomes:

\begin{align}
    \texttt{loss\_rl} &= -\mathbb{E}_{\hat{y} \sim P_{\theta}} [r(\hat{y})],
\end{align}
where $\hat{y} = [\hat{y_1}, ... \hat{y_T}]$ is the sequence generated by the model (transformer, GNN, etc), $P_{\theta}$ and $r(\hat{y})$ is the reward for that given generated sequence. Then we can estimate the gradient of the non-differentiable reward function as follows:

\begin{align}
    \nabla_{\theta}\texttt{loss\_rl}
    &\approx -\mathbb{E}_{\hat{y} \sim p_{\theta}} [r(\hat{y}) * \nabla_{\theta} log p_{\theta}(\hat{y}|\boldsymbol{x})] \label{rlgrad}\\
    &\approx -\mathbb{E}_{\hat{y} \sim p_{\theta}} [r(\hat{y}) * \sum_{t} \nabla_{\theta} log p_{\theta}(\hat{y_t}|\hat{y_1},...,\hat{y_{t-1}},\boldsymbol{x})], \label{rlgrad}
\end{align}

where $\hat{y}$ is the output sequence generated by the model from the input sequence $\boldsymbol{x}$ and $\hat{y_t}$ is the token predicted in a sequence at time step $t$.

%% file: nmt.tex
\section{Neural Machine Translation: NMT}
\label{section:nmt}
In this section, we provide a mathematical framework for the neural machine translation system. We then describe important papers and ideas that are important for the NMT-based code generation systems. These translation systems generally follow a mathematical framework described below. Mostly, they assume to have a parallel corpus for training the model which contains paired input and output sequences. Let the input be $\boldsymbol{x}$ which contains a sequence of tokens $x_1, x_2, x_3, ..., x_n$. Let the ground truth target for the input sequence $\boldsymbol{x}$ be $\boldsymbol{y}$ which contains a sequence of output tokens $y_1, y_2, y_3, ..., y_m$. Let the translation framework be parameterized by $\theta$. Then the overall objective is to maximize the conditional probability of $\boldsymbol{y}$ given $\boldsymbol{x}$. Where $\hat{y_i}$ is the prediction of the token at time-step $i$ and $y_i$ is the ground truth token at time-step $i$. As we can see from the equation below, maximizing this conditional probability is equivalent to minimizing the sum of cross-entropy loss at each time-step that can be minimized using backpropagation through time (BPTT)~\cite{werbos1990backpropagation}. For the language translation tasks $\boldsymbol{x}$ is the input sentence in one language and $\boldsymbol{y}$ is the output sentence in another language. For code generation $\boldsymbol{y}$ is the source code, while $\boldsymbol{x}$ can be any representation of the source code such as assembly code or the natural explanation of the code. Most of the NMT models differ in how they calculate $P(y_i | y_{<i}, \boldsymbol{x}; \theta)$. We can model $P(y_i | y_{<i}, \boldsymbol{x}; \theta)$ using models such as RNNs, transformers, tree decoders, or GNNs. For NMT models, $y_i$ is the ground truth token at decoding step $i$. For tree decoder and GNN, $y_i$ is the ground truth token at node $i$. Thus, for NMT models, the total loss of one sample is the sum of cross-entropy loss at each time of the decoding step. For the tree decoder and GNN model, the total loss is the sum of cross-entropy loss at each node of the decoding step. 
\begin{align*}
    P(\boldsymbol{y}| \boldsymbol{x}; \theta) &= \argmax_{\theta} P(y_1, y_2, y_3,\ ..., y_m | \boldsymbol{x}; \theta)\\
    &= \argmax_{\theta} P(y_1 | \boldsymbol{x}) * P(y_2 | y_1, \boldsymbol{x}; \theta)\
    * ... * P(y_m | y_{m-1}, ... y_1, \boldsymbol{x}; \theta) \quad \because \texttt{chain rule}\\
    &= \argmax_{\theta} \prod_{i=1}^m P(y_i | y_{<i}, \boldsymbol{x}; \theta) \quad \texttt{where: }  y_{<i} = y_1, ..., y_{i-1} \\
    & \texttt{which is equivalent to maximizing the log likelihood}\\
    &= \argmax_{\theta} log \prod_{i=1}^m P(y_i | y_{<i}, \boldsymbol{x}; \theta) \\
    &= \argmax_{\theta}\sum_{i=1}^m log P(y_i | y_{<i}, \boldsymbol{x}; \theta)\\
    & \texttt{which is equivalent to minimizing the negative log likelihood} \\
    &= \argmin_{\theta} -\sum_{i=1}^m log P(y_i | y_{<i}, \boldsymbol{x}; \theta)\\
    & \texttt{let } P(y_i | y_{<i}, \boldsymbol{x}; \theta) \sim \texttt{Categorical(p)}\\ 
    & \texttt{then, using the pmf of a categorical distribution}\\
    &= \argmin_{\theta} -\sum_{i=1}^m log \prod_{k=1}^K p_{k}^\mathbbm{I(\hat{y}=k)}\\
    &= \argmin_{\theta} \sum_{i=1}^m  \sum_{k=1}^K -log p_{k}^\mathbbm{I(\hat{y}=k)}\\
    &= \argmin_{\theta} \sum_{i=1}^m -y_{i} * log p(\hat{y_i})
\end{align*}
Here, \texttt{Categorical(p)} is the categorical distribution over the output vocabulary. All of these models try to maximize the likelihood of the next token given tokens generated till now. Thus, the generated tokens may not be syntactically correct. Some of the neurosymbolic methods often put a constraint on the grammar of the output language at each token prediction step, such that the output is always syntactically correct~\cite{yin2017syntactic}.In the following section, we summarize important papers that build the foundation for current NMT techniques for source code generation. 

\subsection{Sequence to Sequence Learning with Neural Networks}
Sequence-to-sequence networks (Encoder-Decoder architecture) proposed by \cite{sutskever2014sequence} is one of the most widely used frameworks for neural machine translation. The following figure shows their architecture where $sos$ denotes the start of the sequence token, and $eos$ denotes the end of the sequence token. The $h_t$ denotes the hidden state of the LSTM at time step $t$. EMB-E denotes the embedding layer for the encoder and EMB-D denotes the embedding layer for the decoder. LIN denotes the linear layer that transforms the output hidden state of the decoder into the output word probabilities. The $z$ is the context vector from the final time step of the encoder that captures all the information about the input sequence and is passed to the decoder to make the prediction.

\begin{figure}[H]
    \centering
    \includegraphics[width=\textwidth, height=3in]{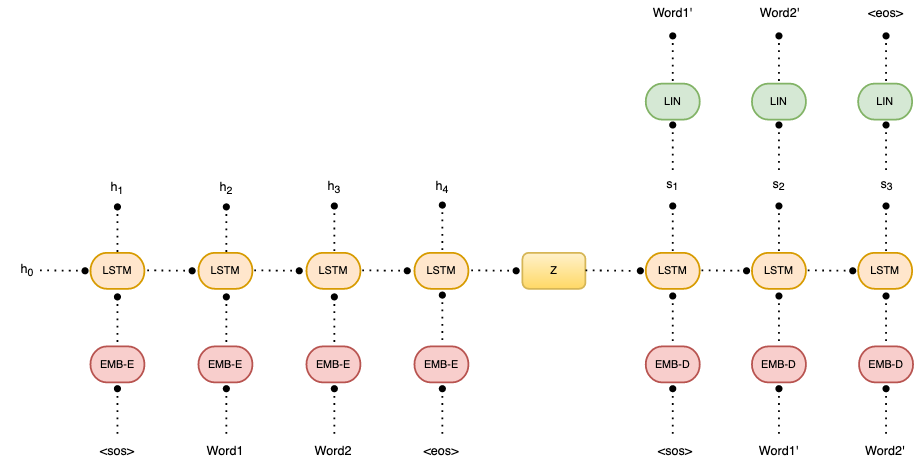}
    \caption{Sequence to sequence learning}
    \label{fig:seq2seq}
\end{figure}

At a single time step, the LSTM takes the embedding vector of input $x_t$ and its previous hidden state $h_{t−1}$ to have its current hidden state $h_t$. The main problem with this mechanism is that it needs to squash all the information about the input sequence into the single context vector $z$, which quickly becomes a bottleneck for longer sequences. The implications for the code generation are that: although this is a good baseline method, it is not that robust at handling long-range dependencies that are frequent in the code generation task.

\subsection{Neural Machine Translation by Jointly Learning To Align And Translate}
The architecture of this sequence to sequence with attention~\cite{bahdanau2014neural} is similar to the above sequence to sequence architecture (encoder-decoder)~\cite{sutskever2014sequence} with the following differences: first, the input token representation (embedding) at each time step ($h_t$) combines the representation (embedding) of a word calculated using the context from both sides of the word. The context from the left side ($h_t^{\rightarrow}$) is calculated using a GRU~\cite{cho2014properties} that takes tokens from the input sequence from the left to the right manner. The context from the right side ($h_t^{\leftarrow}$) is calculated using a GRU~\cite{cho2014properties} that takes tokens from the input sequence from the right to the left manner. Finally, these two representations: ($h_t^{\rightarrow}$) and ($h_t^{\leftarrow}$) are concatenated to get the representation of a word. This helps the model to capture the context of a word from both sides. Second, it reduces the bottleneck problem of previous architecture using an attention mechanism. This architecture lets the model look at the hidden representation of all the time steps of the input sequence while decoding $\hat{y_t}$. Thus the model doesn't need to compress all the information into the context vector $z$. At every decoding step $\hat{y_t}$, this model takes the previous prediction $\hat{y_{t-1}}$ and a vector $c_i$. The context vector $c_i$ is different for every decoding step. First, it calculates the $\alpha$ values, which are the weight (scalar) the model should give to the input word, i.e. $\alpha_{t, 1}$ denotes the weight the model should give to the input word $x_1$ while decoding $y_t$. Then the final context vector $c_i$ is calculated as follows:
\begin{align}
    c_i &= \sum_{j=1}^{T} \alpha_{i, j} h_j
\end{align}
Where $h_j = h_j^{\rightarrow}; h_j^{\leftarrow}$ and $;$ denotes the concatenation operator. They use a multi-layer perceptron (MLP) to calculate the weight $\alpha_{i, j}$. These weights are then converted to a [0, 1] range using a softmax function. The implications for the code generation are that: the attention mechanism helps this model to handle long-range dependencies compared to the previous architecture. \\

\subsection{Attention Is All You Need}
The main problem with the previous two architectures~\cite{sutskever2014sequence, bahdanau2014neural} is that they are sequential, meaning they take one word at a time which increases training time and is hard to parallelize. Transformer~\cite{vaswani2017attention} solves this problem by removing the recurrent neural network architecture that takes one token at a time and using attention layers and fully connected layers that can take the whole sequence at a time. Thus, the model is easy to parallelize across GPUs. Moreover, it overcomes the vanishing gradient and exploding gradient problem of the above two architectures as it does not have long chain multiplication that occurs in backpropagation through time (BPTT). We can see the multi-head attention as generating a contextualized embedding of the input token. As the model takes the whole sequence at a time as the input, it does not preserve the sequential nature of the input and output. They use positional encoding to solve this problem where the embedding of the input word is added to the embedding of the position it occurs in. The implication for the code generation is that: it can handle long-term dependencies more easily than the previous two architectures because each input token is connected to another through a multi-head attention unit. The major disadvantage of this model is that the run time complexity and memory complexity of the system grows quadratic with respect to input length. This can be a big issue for code generation as the input to the code generation (such as assembly) can be quite huge. In such cases, methods that improve this complexity to linear such as BigBird~\cite{zaheer2020big}, and LongFormer~\cite{beltagy2020longformer} can provide a good alternative solution. Moreover, making the attention mechanism IO aware and sparse~\cite{dao2022flashattention} is another alternative direction for handling long-range sequences.

Following table~\ref{table:comparison} shows complexity analysis for a single decoding step of the decoder with respect to the length of the input sequence (n). 
\begin{table}[H]
\caption{Comparison between NMT Models}
\begin{center}
\scalebox{1}{
\begin{tabular}{ ||c|c|c|c|c|c|| } 
 \hline
 Model & run-time complexity & memory complexity & path length & easy to parallelize \\
 \hline
 \seqseq & $O(n)$ & $O(1)$ & $O(n)$ & $\times$\\
 \hline
 \attention & $O(n)$ & $O(n)$ & $O(n)$ & $\times$\\
 \hline
 \transformer & $O(n^2)$ & $O(n^2)$ & $O(1)$ & $\checkmark$\\
 \hline
\end{tabular}}
\label{table:comparison}
\end{center}
\end{table}

Moreover, state space models~\cite{gu2020hippo, gu2021combining, gu2021efficiently, dao2022hungry} are getting popular at handling long-range dependencies better than transformer architecture and exploring this direction for code generation is an interesting and exciting research direction.

\subsection{Towards String-to-Tree Neural Machine Translation}
This paper~\cite{aharoni2017towards} injects syntactic knowledge about the target language into the model by translating the target sequence into a constituency tree.  They linearize this tree into a sequence and use sequence to sequence with attention~\cite{bahdanau2014neural} to generate the target sequence. They show that this improves the prediction and alignment between the input sequence and target sequence compared to direct output sequence prediction without syntactic knowledge. The implications of this for code generation are that: we can convert the source code to an abstract syntax tree (AST) that contains syntactic information and try to predict this AST instead of the source code to get a better model. Direct tree prediction is hard to parallelize, but the linearization step allows us to use the sequence to sequence models such as transformers~\cite{vaswani2017attention} to make a prediction, and it is easy to parallelize. Moreover, there are an increasing number of papers that try to inject syntactic knowledge by pretraining on ASTs~\cite{wang2021syncobert}, and semantic knowledge by pretraining on data flow graphs~\cite{guo2020graphcodebert}. The model that can capture the syntactic structure and semantic structure of the code is better than the one that doesn't. 

\subsection{Tokenization}
Tokenization of the source code is an important component of neural machine translation for source code generation. Developers come up with new words all the time for naming their variables, and functions, thus increasing the output vocabulary size. The complexity of training and decoding the NMT-based system increases with output vocabulary size~\cite{jean2014using}. The tokenization of the source code or natural language can be grouped into the following high-level groups:

\subsubsection{word-based tokenization:}
Word-based tokenization tokenizes the source code or the natural language based on the complete words for frequent words and uses the "<unk>" (unknown) token for rare words. But this degrades the quality of translation if the number of unknown words increases~\cite{bahdanau2014neural}. Given that the source code has really long tail distribution~\cite{le2020deep}, it is important to take care of these tokens. There are multiple solutions to this problem in the literature, for example,~\cite {luong2014addressing} perform alignment of the "<unk>" token with the input sequence and replaces the "<unk>" token with the aligned word from the input sequence. In the case of source code, another possible solution is to replace variables with specific tokens like "var0, var1, ..." (and the same for functions). This is a good solution to reduce the size of output vocabulary, but it makes the generated code unreadable. In summary, this word-based tokenization increases vocabulary size and is problematic for open vocabulary learning. An increase in vocabulary size increases time complexity and memory requirement for NMT-based solutions.

\subsubsection{Character-based tokenization:}
Character level encoding~\cite{chung2016character, ling2015character} uses a single character as a token. The advantage of the character level tokenization is that the output vocabulary is small, but it increases the sequence length by a large amount that is hard to predict for an NMT model. Moreover, it is hard to learn the semantic representation of a character ('c') compared to a word ('computer'). Thus, this is good for reducing the vocabulary size but it makes the model harder to converge as it increases the sequence length by a large amount.

\subsubsection{Subword-based tokenization:}
A hybrid solution between character-level tokenization and word-level tokenization is sub-word-based tokenization. It is based on the idea that frequent words should be tokenized as words and rare words should be tokenized as meaningful sub-words. This is interesting in the case of source code because developers often generate new names using camel-case and snake-case which are quite rare and this technique allows us to split these new words into constituent meaningful sub-words. This allows the sub-word-based tokenization to have a smaller vocabulary and still can learn meaningful representations. There are multiple ways of doing sub-word tokenization such as  byte pair encoding (BPE)~\cite{sennrich2015neural}, WordPiece~\cite{schuster2012japanese}, and  SentencePiece~\cite{kudo2018sentencepiece}. Also, source code contains large numerical values and tokenizing each number separately is not possible. In such cases, it is better to tokenize these numbers into a sequence of digits~\cite{katz2018using}. 

We refer to~\cite{mielke2021between, sun2014empirical, webster1992tokenization, rai2021study} for more analysis on tokenization techniques.

\subsection{Data Augmentation:}
Data augmentation is an important technique to improve the generalization of deep neural networks~\cite{shorten2019survey}. Several techniques such as rotation, flip, random crop, etc exist for data augmentation in computer vision domain~\cite{shorten2019survey}. Several techniques such as random addition, random deletion, synonym replacement, etc exist for NLP domain~\cite{feng2021survey}. We can apply some of the data augmentation techniques from NLP for source code generation. Recently, \neutron tried data augmentation techniques such as random masking and show impressive performance improvement from data augmentation. Moreover, Exploring Data Augmentation for Code Generation Tasks~\cite{chen2023exploring} experiments with data augmentation techniques for source code generation such as using monolingual data using back translation~\cite{sennrich2015improving, currey2017copied}, improving numeric awareness and shows improvement in the code generation. But, the research on data augmentation techniques for code generation is still nascent and is an important research direction.

\subsection{Decoding strategy}
Most of the NMT-based solutions generate source code in an auto-regressive fashion. The next token is predicted based on the maximum likelihood of the token given the input sequence and already generated tokens. It has been shown that this greedy approach has multiple problems such as degenerate solutions~\cite{welleck2019neural} and lacking semantic consistency~\cite{basu2020mirostat}. The greedy approach is computationally efficient and is the optimal choice for the current time step, but when we generate the full sequence, it may be a sub-optimal choice. Another option would be to store every prediction at every decoding step and explore all possible sequences (exhaustive search) from these predictions, this will give us the correct sequence, but this will lead us to an exponential algorithm. The beam search decoding stands in the middle of these two extremes by keeping a fixed number of sequences at each time step based on their joint probability. Thus, it provides a good compromise between accuracy and computational cost. The implication of this for code generation is that: beam search is a powerful method that can improve the accuracy of the source code generation models. Beam search keeps a fixed (beam size) number of predictions at each time step, but beam search with adaptive size could even improve results even better~\cite{freitag2017beam}. Moreover, beam search can be extended to generate syntactically correct source code with some computational cost. These maximization-based methods (greedy and beam search) have been shown to generate solutions with undesirable repetition~\cite{fan2018hierarchical} in the field of NLP. Recently,~\cite{ellis2019write} have shown that sequential monte carlo (SMC) with some value function performs better than beam search for the graphics program generation from images. Sequential monte carlo keeps $K$ number of particles (programs) and re-weights them on some value function. One disadvantage of SMC is that it is more computationally expensive than beam search. Another set of methods that are frequently used for decoding is stochastic methods like top-k sampling~\cite{fan2018hierarchical}, and nucleus sampling~\cite{holtzman2019curious}. These stochastic methods are shown to produce semantically inconsistent text with prefix~\cite{su2022contrastive} in the field of NLP. Another line of decoding strategy is to use another model as a ranker. For example, LEVER: Learning to Verify Language-to-Code Generation with Execution~\cite{ni2023lever} reranks the generated programs based on LM based ranker or verifier that takes the input description, sampled code and execution results and shows improved results. Coder Reviewer Reranking for Code Generation~\cite{zhang2022coder} samples multiple solutions using temperature parameter and uses two models: coder model that evaluates $p(y|x)$ and reviewer mode that evaluates $p(x|y)$ and uses the product of these two scores to select the generated program. They show impressive results and claim state-of-the-art results. Recently, contrastive search-based decoding strategies~\cite{su2022contrastive, su2022contrastivesecond} have shown to be a good decoding strategy for neural text generation. But, their research on source code generation is limited and is an important research direction. NeuroLogic Decoding~\cite{lu2020neurologic} introduces logical constraints during decoding for text generation. NeuroLogic $A^*$~\cite{lu2021neurologic} decoding extends neurologic decoding with lookahead heuristics to generate a text that satisfies the constraints. Similar techniques may be useful for code generation under constraints. Recently, Planning With Large Language Models for Code Generation~\cite{zhang2023planning} propose planning guided transformer decoding where the idea is to use pretrained model to generate complete sequence from a given token (lookahead search) and evaluate it using test cases. This method generated better programs but is computationally expensive.

%% file: sourcecode.tex
\subsection{SourceCode}
These methods generate source code directly from the given input. They mostly use RNN, LSTM, and Transformer techniques for the generation of the source code. The advantage of generating source code directly instead of the AST is that we don't need a module to convert from AST to the source code. Another advantage of direct source code generation is that the same method can be easily transferred for source code generation in multiple languages. The disadvantage of direct source code generation instead of AST generation is that it doesn't contain syntactic information explicitly, making it harder for the model to learn them on its own, and also the representations learned are more specific to the target language~\cite{alon2019code2vec, alon2018code2seq}. In the following section, we briefly summarize papers that generate source code based on the methods they use. 

\subsubsection{RNNs}:
These methods generate source code using recurrent neural networks (RNNs) and their variants such as LSTM and GRU. The problem with RNN-based models is that they can not handle long-range dependencies quite well. For example, \rnndecompilation~\cite{katz2018using} generates C source code from binary sequence using RNNs. They tokenize input binary sequences at the byte level. They do not use existing disassemblers that are good at translating from input binary to assembly, which can give good inductive bias for the neural network as it contains richer semantics and structural information than binary sequence. \neuraldecompilation~\cite{katz2019towards} uses LSTM with attention to generate source code (C) from input assembly representation. They use compiler feedback to retrain the system with some additional data if most of the generated code does not compile. This idea of using compiler feedback for improving the system (either by retraining as done here or using them as reward signals in RL-based systems) is an important research direction for compilable source code generation. \neutron:~\cite{liang2021neutron} uses LSTM with attention model~\cite{luong2015effective} to generate source code (C) from assembly language. They first segment input assembly into different blocks using an LSTM encoder and translate each unit thus generating code fragments in a high-level language. They use data flow, and control flow analysis of the input assembly to compose these code fragments into a function. Moreover, they use feedback from the syntax checker and improve the system further by retraining and using rule-based checkers as error correction techniques. \lpn: Latent Predictor Networks for Code Generation~\cite{ling2016latent} tackles the problem of source code generation in high-level languages like Python and Java from input descriptions using a modified LSTM with attention~\cite{bahdanau2014neural} mechanism. They propose a hybrid approach based on pointer networks~\cite{vinyals2015pointer} and character RNNs (predictors). The model first selects the predictor (latent) conditioned on input and uses that to either generate a character or copy from the input. Thus the same network can generate characters and copy from the input sequence at the same time. \synfix: Automated correction for syntax errors in programming assignments using recurrent neural networks~\cite{bhatia2016automated} uses RNNs to fix syntax errors (\eg missing bracket) in massive open online courses (MOOCs) for buggy code using a model learned from a correct submission for the given problem. They use a sequence model to predict token replacement or insertion at the location provided by the compiler. As the compiler does not always provide the correct location of the bug, \deepfix: Fixing Common C Language Errors by Deep Learning~\cite{gupta2017deepfix} propose RNNs with attention~\cite{bahdanau2014neural} to predict the buggy line and the code to replace it. They use compiler feedback to iteratively fix errors in the program until they are fixed completely. These methods~\cite{bhatia2016automated, gupta2017deepfix} mostly fix syntax-related bugs but trying to solve semantics-related bugs (\eg replacing variables) is also an interesting direction~\cite{devlin2017semantic}. An Empirical Study on Learning Bug-Fixing Patches in the
Wild via Neural Machine Translation~\cite{tufano2019empirical} performs a large-scale empirical study on the viability of these RNN-based NMT models for fixing real-world bugs and comes up with a positive answer. \pixtwocode: Generating Code from a Graphical User Interface Screenshot~\cite{beltramelli2018pix2code} generate source code from graphical user interface. They use CNNs to process the input GUI and generate source code using LSTM. \sparsepointernetwork: Learning Python Code Suggestion with a Sparse Pointer Network~\cite{bhoopchand2016learning} use LSTM with attention to generate new tokens for code completion. They also apply pointer network~\cite{vinyals2015pointer} with attention to existing identifiers to select them for code completion. Thus, the model can generate new tokens and also can select one from the existing set for completion. \nltwobash: A Corpus and Semantic Parser for
Natural Language Interface to the Linux Operating System~\cite{lin2018nl2bash} and Program Synthesis from Natural Language Using Recurrent Neural Networks~\cite{lin2017program} use RNNs with attention to generate bash scripts from natural language explanations. A common limitation of RNN-based code generation is that the performance of these RNN-based decoders decreases as the length of the sequence increases.

\subsubsection{Transformer and LLMs}:
Code generation using transformer~\cite{vaswani2017attention} and large language models~\cite{radford2018improving, brown2020language, devlin2018bert} is the current state-of-the-art method. \codex~\cite{chen2021evaluating} finetunes a language model called GPT~\cite{radford2018improving} on public source code available in GitHub for source code generation from natural language explanation. They generate multiple samples and select the best one based on criteria such as unit tests or mean log probability, and this improves the performance of the system by a large margin. The limitation of this model is that the model performance degrades when the sequence length increases. \codex has also been deployed in real-world toolkits such as GitHub Copilot to assist developers by generating source code from the natural descriptions. \alphacode~\cite{li2022competition} presents a solution that can generate competition-level source code (Python and C++) from natural language examples and unit tests. They pre-train their model on multiple programming languages using masked language modeling like BERT~\cite{devlin2018bert} for the encoder and next token prediction for the decoder~\cite{li2022competition}. They then fine-tune their model on the CodeContests dataset. During inference, they generate millions of samples per problem in parallel using high temperature (\eg T=0.25). Once they generate multiple samples, they filter those codes that can pass the given example unit tests. They also cluster the solutions based on their response to input test cases. These sampling, filtering, and clustering are important steps for the competition-level source code generation~\cite{li2022competition}. \llm: Program synthesis with large language models~\cite{austin2021program} performs a large-scale study of large language models (LLMs) for source code generation (Python) from natural language description and unit tests. They show that these LLMs are quite effective at few-shot learning tasks on code generation. \incoder: A Generative Model for Code Infilling and Synthesis~\cite{fried2022incoder} presents a unified framework for code generation and code editing based on LLMs. They train autoregressive models to predict a masked document (a document where some span is replaced by <mask> and moved to the end). This allows them to generate code at test time and also allows code editing, variable re-naming, type prediction, and comment generation using a single model. \polycoder A systematic evaluation of large language models of code~\cite{xu2022systematic} presents systematic evaluation of autoregressive LLMs such as Codex~\cite{chen2021evaluating}, GPT-Neo~\cite{black2021gpt}, GPT-J~\cite{wang2021gpt}, GPT-NeoX~\cite{black2022gpt}, CodeParrot~\cite{tunstall2022natural} across various programming languages. They also propose a medium-sized LLM called PolyCoder trained in multiple programming languages. They observe the performance of LLMs increases by training using multiple languages, increasing the model size (except CodeParrot), and training longer. \nltwocode: A Scalable and Extensible Approach to Benchmarking NL2Code for 18 Programming Languages~\cite{cassano2022scalable} present parallel code generation benchmark on multiple languages and evaluate two popular LLM-based models \codex and \incoder for code generation. Generating Bug-Fixes Using Pretrained Transformers~\cite{drain2021generating} show that transformr~\cite{vaswani2017attention} based models are also quite effective at fixing bugs. Learning Autocompletion from Real-World Datasets~\cite{aye2021learning} and IntelliCode Compose: Code Generation using Transformer~\cite{svyatkovskiy2020intellicode} use these transformer-based models for real-world code completion. Learning Performance-Improving Code Edits~\cite{madaan2023learning} uses anguage model to predict optimized code (instead of a code that has time complexity $O(n^2)$, suggest code that does it on $O(n)$
). \conversation: A Conversational Paradigm for Program Synthesis~\cite{nijkamp2022conversational} and The Programmer’s Assistant~\cite{ross2023programmer} presents a conversational approach to source code generation (Python) from natural language explanation using LLM based model where the model interacts with a user to generate subprograms multiple times leading to the complete solution. They concatenate the previous prompts with previous subprograms for the next subprogram prediction using LLMs. Making the system capable of generating source code on a conversational paradigm is an important research direction. This allows the model to understand the user intent in a better way. Understanding user intent for code generation is an important first step towards code generation~\cite{gottschlich2018three}. Conversational Automated Program Repair~\cite{xia2023conversational} uses conversational paradigm to fix a buggy code using LLM models like codex and ChatGPT~\cite{ouyang2022training}. Recently, prompt-based engineering~\cite{weng2023prompt, wei2022chain} has been quite popular as a way to understand user intent for natural text generation. Evaluating the Text-to-SQL Capabilities of Large Language Models~\cite{rajkumar2022evaluating} uses prompt-based evaluation of codex on text-to-SQL conversion and shows improved results. DocPrompting~\cite{zhou2022docprompting} use prompting to gather knowledge from changing documentation and APIs (of library functions) to code generation. These retrieval augmentation-based language models are good for incorporating knowledge from constantly changing documentation and APIs. Asking Clarification Questions for Code Generation in General-Purpose Programming Language~\cite{li2022asking} uses clarification questions (CQs) to resolve ambiguities in the user’s intent understanding by using LLM as a CQ generator and code generator. This model is quite interesting as it can ask questions to users when the natural description is ambiguous. Another important paradigm for code generation is the breakdown of a big problem into subproblems and generating programs to tackle each subproblem separately. This is closely related to how humans generate programs. Parsel: A (De-)compositional Framework for Algorithmic Reasoning with Language Models~\cite{zelikman2023parsel} converts a natural description into parsel functions (subproblems) with constraints from human input, then they solve these subproblems with actual implementations using a language model (Codex) and also use a solver to satisfy the constraints. It would be really interesting if the model can decompose a problem into subproblems. One method that tries this idea is Self-planning Code Generation with Large Language Model~\cite{jiang2023self}. It uses LLMs for self-planning. The LLM generates plans for solving a task using prompt engineering and solves the problem in a step-by-step fashion. Another closely related paradigm for code generation is sketch generation. We can think of sketch as a high-level guideline to solve program generation. This idea is also inspired by how humans write source code. We first create a rough sketch and then fill in the small details. Coarse-to-Fine Decoding for Neural Semantic Parsing~\cite{dong2018coarse} first decodes the sketch of the program (with coarse tokens) and secondly decodes the low-level details like variable names based on the input and the sketch. A similar idea has been used in ~\cite{nye2019learning}. SKCODER: A Sketch-based Approach for Automatic Code Generation~\cite{li2023skcoder} combines the idea of retrieval augmented model with sketch-based generation. Given an NL description it chooses a similar code snippet from a retrieval corpus, based on the NL description it uses a sketcher to extract a code sketch from a similar code, and employs an editor to edit the sketch based on the NL description and obtain the target code. Recently, these transformer-based models have also shown their impressive performance in solving math problems by framing them as a program synthesis problem from natural descriptions~\cite{tang2021solving}. Moreover, LLM-based program synthesis is being used to generate programs that can operate on images and natural languages to generate an answer. This mechanism provides a neuro-symbolic framework for solutions to problems in computer vision and natural language processing. For example, PAL: Program-aided Language Models~\cite{gao2022pal} uses code generation as an intermediate step for solving symbolic and arithmetic reasoning. Code as Policies: Language Model Programs for Embodied Control
~\cite{wei2022chain} uses code generation as an intermediate step that can be executed as a policy for robot manipulation. Binding Language Models in Symbolic Languages~\cite{cheng2022binding} takes an input, generates a program (binder program: like a python with pandas) using a language model (Codex) that can then be executed by the binder interpreter (SQL + python + language model) to produce an answer for natural language question-answering tasks. The performance of these LLMs increases with the increase in their model size~\cite{austin2021program, chen2021evaluating}, but the large size makes them expensive to train and evaluate.

\subsubsection{RL}:
Code generation models based on RL can utilize the signals from unit tests or compilers to improve the code generation system unlike previous RNNs and LLMs-based models (\alphacode uses unit tests to filter out samples but doesn't learn from its feedback). For example, \coderl~\cite{le2022coderl} uses actor-critic framework~\cite{konda1999actor, sutton1999policy} where the actor (LLM~\cite{wang2021codet5}) generates a code and a critic (transformer) generates feedback signals for the actor. Similarly, Execution-based Code Generation using Deep Reinforcement Learning~\cite{shojaee2023execution} executes the generated code and uses the results from the execution as rewards to update the model along with KL divergence penalty to reduce memorization, AST, and DFG matching scores for syntactic and semantic knowledge. \repl: Write, Execute, Assess: Program Synthesis with a REPL~\cite{ellis2019write} use RL-based learning paradigm to utilize the signals from REPL(read-eval-print-loop) to generate the source code (graphic programs) from images. This paper is based on the idea of execution-guided program synthesis~\cite{chen2018execution, zohar2018automatic} where the idea is to use the execution states of the current subprogram to generate a better program. This is a powerful paradigm that can be combined with any encoder-decoder architecture~\cite{chen2018execution}. This paper uses a policy network ($\pi$) to select an action (production rule from the grammar: add circle/add rectangle rule for 2D graphics) and create a subprogram, REPL to execute the current subprogram and generate current output (render the current program to generate an image), and a value function ($v$) to assess how likely the current program will help towards the final goal (spec: final image to be rendered). Another interesting part of this paper is the evaluation framework where they generate programs using sequential monte carlo~\cite{doucet2001introduction} guided by the value function ($v$) and show that it performs better than the beam search. This is similar to how humans write programs: write something, evaluate it, and correct it until it works. This line of work is a good research direction. Similarly, \seqtwosql: Generating Structured Queries from Natural Language using Reinforcement Learning~\cite{zhong2017seq2sql} uses a policy network to generate SQL queries from natural language descriptions and utilize the signals from the execution of a generated query. The reinforcement framework allows us to train a system where we can have multiple SQL queries producing the same result using rewards, which are otherwise penalized by cross-entropy loss except for the ground truth query. \rlcorrection: Deep reinforcement learning for programming language correction uses actor-critic framework~\cite{mnih2016asynchronous} to fix common errors in C source code. They show that the system can be trained faster with expert demonstrations and eventually beat \deepfix. \compcoder: Compilable Neural Code Generation with Compiler Feedback~\cite{wang2022compilable} use compiler feedback as a reward signal to improve the LLM for source code (Python) completion and generation from natural language description. Thus, LLMs with RL framework are quite powerful models for code generation.

\subsubsection{NeuroSymbolic}:
Neurosymbolic methods~\cite{garcez_neurosymbolic_2023} contain a neural module and a symbolic module. Neural networks are great at learning from noisy data but are not interpretable and lack explicit reasoning capabilities. Symbolic models are great at reasoning but are quite inefficient at learning from noisy data. Thus they complement each other for learning from noisy data with reasoning. Thus, neurosymbolic methods are getting quite popular. In this section, we will cover neurosymbolic methods for program generation. We also include library learning methods (methods that build a learned library) inside neurosymbolic methods treating the library as a symbolic module. In the context of code generation, there are two primary ideas: one is to use classical search-based techniques and another is to use neural networks. Classical techniques enumerate over the program space such as enumerative search, or add some heuristics and constraints to speed up such as sketch-based solutions and SMT solver-based solutions. The major disadvantage of these methods is we need to perform a combinatorial search in the program space which is really hard and these models only work with DSLs (do not scale to real-world programming languages like Python and C++). Neural network-based methods generate source code based by predicting the next likely token each time and thus can fail if one token is wrong in the sequence. Methods like DeepCoder: Learning to Write Programs~\cite{balog2016deepcoder} use both techniques for code generation. It uses a neural network to predict the presence or absence of high-level functions (e.g. sort, max, min, reverse) based on input-output examples and let this result guide the search process of classical program induction techniques such as sketch-based solution~\cite{solar2009sketching} and $\lambda^2$~\cite{feser2015synthesizing}. They show that their solution improves the search time by a large margin. The limitation of the paper is that the solution has been tried for a small DSL and applying a similar idea for source code generation in real-world programming languages such as Python, C, and C++ is an interesting research direction. Library Learning for Neurally-Guided Bayesian Program Induction ~\cite{ellis2018library} and Dreamcoder~\cite{ellis2020dreamcoder} proposes a program induction algorithm that learns DSL (higher abstractions) while jointly training a neural network to predict program properties like DeepCoder. This is related to the programming paradigm where a programmer builds libraries of reusable subroutines that are shared across related programming tasks and can be composed to generate increasingly complex and powerful subroutines. The system extracts new DSL subroutines from a common structure found across syntax trees of the generated programs that solve the given set of tasks. These methods are promising directions for human-like code generation but their applicability to source code generation in real-world programming languages such as C, C++, and Python is yet to be seen as the search space of these real-world programming languages is extremely huge compared to the search space defined by the DSL these methods use.

%% file: astsequence.tex
\subsection{ASTSequence}
These papers first represent the program in the form of AST, but at prediction time they linearize the AST into a sequence and make a prediction. The advantage of such a system is that the linearized system is easy to parallelize and the AST representation carries explicit syntax information helping the model learn the grammar of the language easily. The disadvantage of this method is that we need to introduce new tokens (\eg brackets) for linearization of the AST that introduce long-range dependencies in the prediction. Thus, if we miss one corresponding token, the output can't be converted into AST using automated code and thus we can't generate the source code. But these constraints can be satisfied during decoding. For example, \pointermixture: Code completion with neural attention and pointer networks~\cite{li2017code} use this framework for code completion in a dynamically-typed language like Python. They propose a Pointer mixture network where the model can generate new tokens from the output vocabulary using LSTM and attention~\cite{bahdanau2014neural}, and also can copy tokens from the partial code by using pointer networks~\cite{vinyals2015pointer}. They use in-order depth-first traversal to flatten the AST representation of the source code. They also propose parent attention for AST-based code completion where the idea is that the parent node should have high relevance for the child node compared to other nodes. Representing source code using AST and letting the model predict AST instead of source code makes the model easier to converge as the syntax is explicit. The model also learns representations that can be shared across multiple languages~\cite{alon2019code2vec, alon2018code2seq} compared to the model that directly predicts the source code. But, converting source code to AST and flattening them increases the size and introduces long-range dependencies between tokens~\cite{zhang2019novel}. The deeper ASTs also weaken the capability of the models to capture complex semantics~\cite{zhu2015long}. The solution to this problem is to augment the AST with data flow and control flow structures~\cite{allamanis2017learning}. Another solution would be to decompose large ASTs into a sequence of small statement trees~\cite{zhang2019novel}. The methods that use ASTs for injecting syntacting structure into the model and data flow graphs for injecting semantic awareness to the model are rising. Also, comparing models based on multiple output representations and the impact of output representation on code generation is an important research direction.

%% file: binarytree.tex
\subsubsection{BinaryTree}
These methods first encode the Nary AST tree into a binary tree using predefined algorithm like the left child right sibling algorithm and generate a binary tree. The disadvantage of this idea is that we again need another module that converts from binary tree to Nary AST tree. The conversion from an AST tree to a binary tree also increases the long-range dependency between tokens and it can make the training harder. Neural Code Completion~~\cite{liu2016neural} convert the source code to AST. They then convert the AST to a binary tree using the left-child right-sibling algorithm. They use LSTM to predict the next node for code completion. \treetotree~\cite{chen2018tree} translates programs written in one language (Java/Coffeescript) to another language (C\#/Javascript). They generate the AST of the target language given the AST of the input language using a tree decoder. They use tree-lstm~\cite{tai2015improved} to encode the input AST representation of the source code. When the decoder expands a non-terminal, it locates the corresponding sub-tree in the source tree using an attention mechanism and uses the information of the subtree to guide the non-terminal expansion. They also use the idea of parent attention feeding (if target node t depends on source node s, it’s likely that the child of t depends on the child of s ) to improve the prediction results. \coda: An End-to-End Neural Program Decompiler~\cite{fu2019coda} generates AST representation of the source code from the input assembly. They use instruction type aware encoders (separate RNNs for different types of statements instruction types: memory, arithmetic, and branch operations) to encode the input assembly. They generate a binary AST tree by converting the original AST into a binary tree using left child right sibling representation and use  an AST tree decoder with an attention mechanism for decoding similar to \treetotree. The tree decoder consists of two LSTMs, one for the left-child prediction and the another for the right-child prediction.

%% file: narytree.tex
\subsubsection{NaryTree}
These methods directly predict the Nary AST tree. The advantage of this idea is that we don't need another module that converts from a binary tree to a Nary AST tree compared to binary tree generation. The disadvantage of such methods is that it's non-trivial to parallelize these models compared to sequence-to-sequence models. In the following section, we divide the Nary AST tree prediction methods based on the methods they use such as RNNs, transformers, GNNs, and neurosymbolic.

[RNNs and CNNs]: \snm: A syntactic neural model for general-purpose code generation~\cite{yin2017syntactic}  generates abstract syntax trees and then uses them to generate high-level programming language code like Python from the natural language description of the source code. They use the encoder-decoder paradigm where the encoder is a simple LSTM network that encodes the description of the code. The decoder generates ASTs from the encoded input and uses the grammar of the AST. At each step of the generation, it selects one production rule (function call, if-statement) and adds it to the partial AST if the generation step is non-terminal. If the generation step corresponds to the terminal, it generates variable names and values using a copy mechanism using pointer networks~\cite{vinyals2015pointer}. They also use parent feeding while generating a subtree (passing the embeddings of the parent that generated the children and using that for prediction). \lcpc: Mapping Language to Code in Programmatic Context~\cite{iyer2018mapping} uses encoder-decoder architecture to transform the natural language description of method names to generate the code for that method. This paper selects rules using the attention mechanism and uses those rules for decoding the program structure (ATSs). They also use a supervised copy mechanism from using CopyNet~\cite{gu2016incorporating}. \asn: Abstract Syntax Networks for Code Generation and Semantic Parsing~\cite{rabinovich2017abstract} propose abstract syntax networks (ASNs) that generate source code abstract syntax trees [ASTs] (Python) in a top-down manner from natural descriptions extracted from card games such as HearthStone. They use an encoder-decoder mechanism where the encoder encodes the natural language [extracted from images] and the decoder constructs ASTs for the source code in a top-down manner. They use vertical and horizontal LSTMs with attention to transferring information in the top-down and horizontal directions respectively. They use different modules corresponding to the construct in the grammar. The advantage is that the network always respects the grammar of the language being generated. The network first predicts the module and the corresponding module does the expansion. They also use the supervised loss for the alignment (attention) of target tokens with some input tokens instead of using attention as a post-processing step for the copy mechanism. Program Synthesis and Semantic Parsing with Learned Code Idioms~\cite{shin2019program}  mines code idioms (fragments of code that represent higher level abstraction: sum of two numbers) from the dataset by using their ASTs and looking at the repetitive patterns. During the program synthesis, it generates an AST: but for each node generation, it can either generate a token or a code idiom (subgraph of AST). This code idiom mining is an important idea for building subroutines that can be used across multiple tasks for source code generation. \cnndecoder A grammar-based structural CNN decoder for code generation~\cite{sun2019grammar} use convolutional neural networks (CNNs)~\cite{lecun1995convolutional} instead of RNNs for predicting the grammar rules in a similar fashion of \asn. They show that CNNs can handle long-range dependencies better than RNNs.

[Transformer and LLMs]:
\slm: structural language models~\cite{alon2020structural} uses abstract syntax trees (ASTs) for code completion. Given the source code of the program with some lines left out, the model can predict the left-out code. The main idea is to jointly learn the encoder and decoder. Given the source and target both are codes, they jointly learn the probability of the program’s AST. The probability of an AST is calculated as a conditional probability of each node given other nodes observed so far. To gather the information of each node for predicting what the next node will be, they use paths from the root to the given node and paths from every leaf terminal to the given node (encoded using LSTM). They then use transformer~\cite{vaswani2017attention} to aggregate information from multiple paths. AST Path-based representation of the source code has been quite powerful for source code representation~\cite{alon2019code2vec, alon2018code2seq}. \treegen: A Tree-Based Transformer Architecture for Code Generation~\cite{sun2020treegen}  uses transformer~\cite{vaswani2017attention} to generate AST representation of the source code by making a prediction over the grammar rules from natural explanation. The advantage of this method is that the transformer can handle long-range dependencies well compared to RNNs and the prediction over the grammar rule makes sure that the generated code is syntactically correct.
   
[GNNs]:
\gmg: Generative Code Modeling with Graphs~\cite{brockschmidt2018generative} attempts to do code completion given some context code using graph neural networks. The main idea is to use ASTs to represent the given partial code and augment it using some type of edges between nodes in partial ASTs like the parent of edge, child of edge, next sibling edge, etc (data flow and control flow augmentation). They then apply a graph neural network to get the representation of the partial program. During each expansion of the node, they perform classification on the grammar rules and select one among the possible grammar rules so that the subtrees generated are always syntactically correct. 

[NeuroSymbolic:]
\semfix: Semantic Code Repair using Neuro-Symbolic Transformation Networks~\cite{devlin2017semantic} developed a system to predict the location of the simple semantic bugs (incorrect variable, incorrect comparison operator, missing not operator, missing self) along with the actual fix without using unit tests for Python. They convert the source code to AST, embed the AST node with information such as the string form of the node, position of the node, relationship with the parent, and the type of the node, and encode AST using bidirectional LSTM. They then use another module to select the repair candidate rules (\eg replace == by != at node n). They use MLP to select repair candidate rules as they are fixed and pointer networks ~\cite{vinyals2015pointer} to variable replacement. The limitation of this paper is that the developers need to write these repair candidate rules. The research on using neurosymbolic methods for code generation is limited and is an important research direction.

%% file: copy.tex
\section{Copy Mechanism}
\label{section:copy}
Copy mechanisms: a system for copying the tokens from input to output are quite useful for code generation. For example, they can be extremely useful for neural decompilation where we need to copy the values from input assembly to output code. They are equally important for code generation from natural language, program translation, and more. There are multiple solutions to this problem. \ptrnet: Pointer Networks~\cite{vinyals2015pointer} proposes the idea of using attention as a pointer to select a member of the input sequence as the output token. This allows the model to select some part of the input at the time of generation. This idea has been quite powerful and multiple papers in machine translation and code generation use this idea or its variants to copy some tokens from the input sequence. It also allows us to limit the output vocabulary during code generation as we can copy some of the tokens from the input directly (even if they are out of vocabulary tokens for the output language). \ptrgen: Get To The Point: Summarization with Pointer-Generator Networks~\cite{see2017get} uses sequence-to-sequence networks for abstract text summarization with the hybrid approach of generation of tokens and copying values from the input using pointer networks. They also use a penalty term in the loss function that penalizes repetitive attention to the same location. \copynet: Incorporating Copying Mechanism in Sequence-to-Sequence Learning~\cite{gu2016incorporating}. It introduces a mechanism to copy values from the input sequence to the output sequence in the sequence-to-sequence learning scenarios. Their method decides when to copy from the input or when to predict the target value using two functions. The function calculates attention between the current hidden state of the decoder to every word in the vocabulary and to every word in the input sequence and decides to copy or generate the target token. Moreover, the system can be trained in an end-to-end fashion from the data. Thus copying mechanism is an important part of the neural machine translation systems for code generation. But, in some cases, direct copying from the input sequence may not be a good option as the output token depends on some function of the input token. For example, when we are trying to generate a source code from an assembly sequence where the multiplication by two in the source code is implemented by shift operations, direct copying may give us the wrong results. In such cases, complementing the copying mechanism with simple functions like multi-layer perceptron could give better results.

%% file: repr.tex
\section{Representation Learning and pretraining}
\label{section:representation}
Learning a good way to represent the source code or the text is an important part of the source code generation. The embeddings learned from these techniques can be used for various downstream tasks such as code generation, code summarization, and code completion. In this section, we summarize various ideas that can be used for representation learning and pretraining. Initial work on code representation is inspired by works in NLP such as \wordtovec: Efficient Estimation of Word Representations in Vector Space and Distributed Representations of Words and Phrases and their Compositionality~\cite{mikolov2013efficient}. It uses a distributional hypothesis that states words that occur in similar contexts have a similar meaning. Pretraining strategies are quite popular in NLP and CV as a way to utilize unlabelled data to learn the representations. In the following sections, we will describe some of the important pretraining and representation learning strategies for source code.
NMT-based code generation uses an encoder-decoder-based model for code generation. Based on the component(encoder, decoder, bot) they use for pertaining, these methods can be divided into the following groups:

\subsection{Encoder only:}
These methods pre-train only the encoder. For example, \cubert: Learning and Evaluating Contextual Embedding of Source Code~\cite{kanade2020learning} uses BERT's~\cite{devlin2018bert} masked language modeling objective for code generation pre-training. \codebert: A Pre-Trained Model for Programming and Natural Languages~\cite{feng2020codebert} uses masked language modeling along with replaced token detection for code generation. \graphcodebert~\cite{guo2020graphcodebert} use data flow information extracted from code in conjunction with \codebert. \dobf~\cite{roziere2021dobf} uses deobfuscation based pre-training to inject programming language domain information. \dobf and \graphcodebert focus only on the code-specific encoder. The disadvantage of these methods is that they do not utilize the decoder for pre-training.

\subsection{Decoder only:}
These methods only pre-train the decoder. For example, \intellicode~\cite{svyatkovskiy2020intellicode} uses \gpt for code completion task and are trained for next-word prediction. The disadvantage of these models is that they do not utilize the encoder during the pre-training stage. 

\subsection{Encoder-Decoder:}
These methods pre-train both encoder and decoder at the same time. For example, \codetfive~\cite{wang2021codet5} augments model T5~\cite{raffel2020exploring} that employs denoising sequence to sequence pre-training (corrupt the input and lets the decoder decode it) and adds identifier tagging, masked identifier prediction, masked span prediction, and dual generation to improve the system. As this model can pre-train both encoder and decoder for their respective tasks, this method has an advantage over encoder-only and decoder-only methods for source code generation~\cite{wang2021codet5}. \plbart~\cite{ahmad2021unified}: trains on with denoising sequence to sequence models like BART~\cite{lewis2019bart} but does not inject domain information from programming languages domain. \langagnostic~\cite{zugner2021language} proposes to capture the relative distances between code tokens over the code.\\ \\
Based on the type of information these methods try to capture, they can be divided into the following groups:
\subsection{text-based}
text-only models of representation learning such as ~\cite{wang2021codet5} capture only the text-based representation of the source code. But, source code has a richer structure that can be defined using abstract syntax trees (ASTs) and semantics that can be captured using data flow graphs (DFGs) and program dependence graphs (PDGs). The text-only representation learning forces the model to learn these syntactical information and semantic information on its own and thus may not be the optimal way for representation learning of the source code.

\subsection{syntax-based}
These methods try to inject the syntactical knowledge of the source code into their representation learning objective. For example, \codetovec: Learning Distributed Representations of Code~\cite{alon2019code2vec} and \codetoseq: Generating Sequences from Structured Representations of Code~\cite{alon2018code2seq} generates a vector representation of the given code snippet by generating a vector representation for AST paths (path between leaf nodes in an AST) and aggregating them using an attention module. These representations can be useful for tasks like code summarization, documentation, retrieval, code completion, and more. \codetoseq uses LSTM to embed paths into a fixed length vector while \codetovec uses a linear layer. Language-Agnostic Representation Learning of Source Code from Structure and Context~\cite{zugner2021language} leverages context (source-code) and structure (AST) to learn a good representation of the source code. UniXcoder: Unified Cross-Modal Pre-training for Code Representation~\cite{guo2022unixcoder}  proposes multimodal pretraining that utilizes code comments and linearized AST along with the source code. It uses masked language modeling, next token prediction, denoising objective along with multi-modal contrastive learning objective (similar embeddings for similar examples and different for different), and cross-modal generation objective (generate comment from AST and vice versa) as pretraining strategies. In a similar way, TreeBERT~\cite{jiang2021treebert} uses the encoder-decoder transformer framework and utilizes the tree structural information by modeling AST paths (tree-based masked language modeling and node order prediction). SYNCOBERT: Syntax-Guided Multi-Modal Contrastive Pre-Training for Code Representation~\cite{wang2021syncobert} incorporates abstract syntax tree for pretraining. It uses Identifier prediction and AST edge prediction as two pretraining strategies and tries to maximize the mutual information between multimodal inputs such as code, comments, and AST using contrastive learning. SPT-Code: Sequence-to-Sequence Pre-Training for Learning Source Code Representations~\cite{niu2022spt} uses both text and linearized AST of a code with an encoder-decoder framework for code representation learning. 

\subsection{semantic-based} 
These methods try to inject not only text-based and syntactic information, but also semantic information into their learning objectives. Given, source code-based representation learning methods capture context and AST-based techniques capture syntax, hybrid methods that try to learn the context and syntax are increasing. For example,  \lrpg: Learning to Represent Programs with Graphs~\cite{allamanis2017learning} augments the AST with data flow and control flow edges. They use graph neural networks over these graphs to learn a semantic representation of the source code. Contrastive code representation learning~\cite{jain2020contrastive}  uses contrastive learning for source code representation learning. The idea is to maximize similarity with equivalent programs and minimize similarity with functionally different programs. They claim that it is better than masked language models (MLM) as they (MLM) only learn local language reasoning which may not be the best way to summarize the functionality of the program. StructCoder: Structure-Aware Transformer for Code Generation~\cite{tipirneni2022structcoder} encodes syntactic and semantic information to the encoder and decoder for code generation using AST path prediction for syntactic information and data flow for semantic information. Recently, Flow2Vec: Value-Flow-Based Precise Code Embedding~\cite{sui2020flow2vec} uses value flows for code representation learning and shows impressive results. 

%% file: open.tex
\section{Open problems and research directions}
\label{section:open}
Source code generation is a challenging task. There are multiple open problems and research directions. In this section, we will try to highlight some of the open problems and potential research directions. Source code has long dependencies in multiple places~\citep{le2020deep}. For example, a statement to open a file can be in line 3, while the command to close the same file can be in line 100. This really long dependency is a problem for existing techniques. It has been shown that the performance of these encoder-decoder models decreases as the sequence length increases. Thus solving long-range dependencies is an important research direction for source code generation. Most of the existing methods work on next-token prediction (NTP) and optimize the likelihood of the next token given previous tokens and input and this leads to accumulating errors~\cite{bengio2015scheduled, ranzato2015sequence}. Current NMT-based systems work on next token prediction in a sequential fashion. Although different decoding strategies help in avoiding these pitfalls, they are computationally expensive. Humans do not generate a complete program token by token for a complete task like current code generation methods~\cite{li2022competition}. We break down a problem into subproblems and test them iteratively. Methods that can break down a problem into small problems, generate code for such subprograms, and evaluate them such as~\cite{ellis2019write, zelikman2023parsel} are good potential research directions. Most of the current code generation systems do not combine the generated code abstractions into higher-level abstractions as humans do. Humans keep a collection of subroutines and combine them to perform higher-level tasks. \dreamcoder: growing generalizable, interpretable knowledge with wake-sleep bayesian program learning~\cite{ellis2020dreamcoder} takes an initial step in this direction. Most of the best-performing methods in current source code generation are based on large language models. But, these LLMs require a lot of training data and compute resources. Thus, the current solutions are not sample efficient. Making these source code generation samples efficient is an interesting research direction. The syntax-guided method takes care of syntax but not semantics. In particular, standard supervised training procedure could suffer from program aliasing: for the same input-output examples, there are multiple semantically equivalent programs, but all except the one provided in the training data will be penalized as wrong programs~\cite{chen2018execution}. To mitigate this~\cite{bunel2018leveraging} propose to train with reinforcement learning so that it rewards all semantically correct programs once they are fully generated. But~\cite{chen2018execution} show we can use execution-guided synthesis for this task. Execution-guided synthesis~\cite{chen2018execution, zohar2018automatic} currently works with DSLs, but extending them to real-world source code generation is also an interesting research direction. Humans don't write everything from scratch for descriptions to code generation: A Retrieve-and-Edit Framework for Predicting Structured Outputs~\cite{hashimoto2018retrieve} retrieves the closest example from the training dataset and uses an editor network to predict edits on the retrieved code so that it does the intended job. Using these ideas of already existing code and documentation to better improve code generation is an interesting direction. Utilizing ideas such as Unsupervised Translation of Programming Languages~\cite{lachaux2020unsupervised} when we don't have a parallel aligned dataset is also an interesting research direction for source code generation. As more and more papers utilize the feedback from unit tests for improving the code generation models, it is worth exploring automatic unit test generators based on deep learning such as Unit Test Case Generation with Transformers and Focal Context~\cite{tufano2020unit}. This allows us to improve both models, i.e. code generator and unit test generator simultaneously. Most of the existing systems work on small programs with single functions and extension to a program with multiple functions is challenging and interesting research directions. The code generation module may be just remembering the training dataset leading to security and copyright issues: WhyGen: Explaining ML-powered Code Generation by Referring to Training Examples~\cite{yan2022whygen} use fingerprints to find the closest training example. But the scalable solution to internet scale training is still an open research problem. Reinforcement learning from human feedback (RLHF) has been quite popular in fine-tuning large language models with human feedback. Applying similar ideas for improving code generation is an important research direction. Recently, Improving Code Generation by Training with Natural Language Feedback~\cite{chen2023improving} shows initial impressive results in this research direction.

%% file: dataset.tex
\section{Datasets}
\label{section:datasets}
In this section, we will introduce some of the datasets for code generation tasks. CodeXGLUE: A Machine Learning Benchmark Dataset for Code Understanding and Generation:~\cite{lu2021codexglue}[\url{https://github.com/microsoft/CodeXGLUE}] consists of 14 datasets for 10 diversified code intelligence tasks covering code-to-code, code-to-text, text-to-code, and text-to-text (documentation) translation. The dataset covers multiple languages such as C, C++, Java, Python, and more. CodeXGLUE also includes eight previously proposed datasets — BigCloneBench~\cite{svajlenko2014towards}, POJ-104~\cite{mou2016convolutional}, Devign~\cite{zhou2019devign}, PY150~\cite{raychev2016probabilistic}, Github Java Corpus~\cite{allamanis2013mining}, Bugs2Fix~\cite{tufano2019empirical}, CONCODE~\cite{iyer2018mapping}, and CodeSearchNet~\cite{husain2019codesearchnet}. APPS: Measuring Coding Challenge Competence With APPS~\cite{hendrycks2021measuring}[\url{https://github.com/hendrycks/apps}] consists of a dataset for Python code generation from natural language explanation. SPoC: Search-based Pseudocode to Code~\cite{kulal2019spoc}[\url{https://github.com/Sumith1896/spoc}] is a dataset for C++ code generation from pseudocode. Concode: Mapping Language to Code in a Programmatic Context~\cite{iyer2018mapping}[\url{https://github.com/sriniiyer/concode}] is a dataset for Java code generation from docstrings. CodeSearchNet Challenge: Evaluating the State of Semantic Code Search~\cite{husain2019codesearchnet}[\url{https://github.com/github/CodeSearchNet}] consists of a dataset for code retrieval from natural language for multiple languages such as Go, Java, JavaScript, PHP, Python, and Ruby. It can also be used for code generation tasks. A parallel corpus of Python functions and documentation strings for automated code documentation and code generation~\cite{barone2017parallel}[\url{https://github.com/EdinburghNLP/code-docstring-corpus}] consists of a dataset for code generation in Python from docstring and vice versa. StaQC: A Systematically Mined Question-Code Dataset from Stack Overflow~\cite{yao2018staqc}[\url{https://github.com/LittleYUYU/StackOverflow-Question-Code-Dataset}] consists of a dataset for question-to-code pairs for python and SQL mined from stack-overflow. CoNaLa: The Code/Natural Language Challenge~\cite{yin2018mining}[\url{https://conala-corpus.github.io/}] consists of a dataset for python code generation from natural language. HumanEval: Hand-Written Evaluation Set~\cite{chen2021evaluating}[\url{https://github.com/openai/human-eval}] consists of a dataset for handwritten Python program generation from function signature, docstring, and body with an average of 7.7 unit tests per program. CodeContests: competitive programming dataset~\cite{li2022competition}[\url{https://github.com/deepmind/code_contests}] includes CodeNet dataset as well \alphacode Competition level dataset for Python and C++ code generation from a natural language with unit tests. MBPP (mostly basic programming problems) dataset~\cite{austin2021program}[\url{https://huggingface.co/datasets/mbpp}] consists of a dataset for Python code generation from natural language explanation along with test cases. Spider: A Large-Scale Human-Labeled Dataset for Complex and Cross-Domain Semantic Parsing and Text-to-SQL Task~\cite{yu2018spider}[\url{https://yale-lily.github.io/spider}] consists of a dataset for SQL code generation from natural language. NL2Bash: Generating bash command from natural language~\cite{lin2018nl2bash}[\url{https://github.com/TellinaTool/nl2bash}] consists of a dataset for bash script generation from natural language explanation. \polycoder: A Systematic evaluation of large language models of code [\url{https://github.com/VHellendoorn/Code-LMs}] consists of a dataset for code generation in 12 programming languages from natural language explanation. Pix2code: Generating Code from a Graphical User Interface Screenshot~\cite{beltramelli2018pix2code}[\url{https://github.com/tonybeltramelli/pix2code}] consists of a dataset for code generation(Android, IOS, Web Technologies) from a graphical user interface. WikiSQL: Generating Structured Queries from Natural Language using Reinforcement Learning~\cite{zhong2017seq2sql}[\url{https://github.com/salesforce/WikiSQL}] consists of a dataset for SQL query generation from a natural explanation. MultiPL-E: A Scalable and Extensible Approach to Benchmarking NL2Code for 18 Programming Languages~\cite{cassano2022scalable}[\url{https://github.com/nuprl/multipl-e}] consists of a dataset for source code generation in 18 programming languages from natural language explanation. CodeNet: A Large-Scale AI for Code Dataset for Learning a Diversity of Coding Tasks~\cite{puri2021codenet}[\url{https://github.com/IBM/Project_CodeNet}] consists of a dataset for multiple tasks like source code generation, code translation, etc in 55 different programming languages. DS-1000: A Natural and Reliable Benchmark for Data Science Code Generation~\cite{lai2022ds}[\url{https://ds1000-code-gen.github.io/}] consists of a dataset for code generation problems in data science from natural descriptions. The Stack: 3 TB of permissively licensed source code~\cite{kocetkov2022stack}[\url{https://huggingface.co/datasets/bigcode/the-stack}] is a 3.1 TB dataset consisting of permissively licensed source code in 30 programming languages. CoderEval: A Benchmark of Pragmatic Code Generation with Generative Pre-trained Models~\cite{yu2023codereval} consists of a dataset for code generation from natural language collected from real-world open-source projects. Multi-Turn Programming Benchmark (MTPB)~\cite{nijkamp2022codegen} consists of a dataset for conversational code generation where the user specifies the subtasks and the model completes them.

%% file: evaluation.tex
\section{Evaluation}
\label{section:evaluation}
In this section, we will introduce common metrics that are being currently used for generated code evaluation.
\begin{itemize}
    \item BLEU Score: BLEU score~\cite{papineni2002bleu} and exact match accuracy. Exact match accuracy is the fraction of test samples that the model predicts the entire sequence correctly. The BLEU score has been a standard for evaluating machine translation against human translation in natural language processing. The BLEU score is defined as follows:
    \begin{align*}
        \texttt{BLEU} &= \texttt{BP}. exp \big( \sum_{n=1}^N w_n logp_n \big)
    \end{align*}
    
    Where, $w_n$ is a scalar such that $\sum_{n=1}^N w_n = 1$, $p_n$ is a modified precision for n-gram that is defined as follows ($\hat{y}$ is the prediction):
    
        \[p_n = \frac{\sum\limits_{n-gram \in \hat{y}} Count_{clip} (n-gram)}{\sum\limits_{n-gram \in \hat{y}} Count (n-gram)}\]
        
    Where, $Count(n-gram)$ is the number of n-grams in the prediction, and $Count_{clip} (n-gram)$ is the number of these n-grams in the prediction that are present in the target sentence but clipped by the maximum frequency of that n-gram in the target sentence. 
    BP is a brevity penalty that penalizes short translations and is defined as follows:
    
    \begin{equation*}
        \texttt{BP} = \begin{cases}
            1, & \text{if}\ c > r \\
            e^{(1-\frac{r}{c})}, & \text{otherwise}
        \end{cases}
    \end{equation*}
    Where $c$ is the length of the prediction and $r$ is the length of the ground truth target (reference). This gives us a value between [0, 1] for every translation. Generally, N is set to 4, and $w_n$ is set to $\frac{1}{N}$. \\
   
    The problem with the bleu score is that it can not measure the functional correctness of the programs and can not capture semantic features specific to code:~\cite{le2022coderl, hendrycks2021measuring, chen2021evaluating, ren2020codebleu}. Also,~\cite{chen2021evaluating} shows some examples where the functionally equivalent codes have lower blue scores than functionally nonequivalent ones.
    This can be explained by the fact that semantically identical programs can potentially have very low n-gram overlap; for example, because of identifier renaming~\cite{austin2021program, hendrycks2021measuring, chen2021evaluating}. Although the BLEU score has been used in multiple papers for evaluation, the BLEU score is not a good metric for the evaluation of the generated source code. Does BLEU Score Work for Code Migration~\cite{tran2019does} shows that BLEU is not a good measure for code evaluation and proposes a method called RUBY based on string edit distance (text: code similarity), tree edit distance (AST: syntactic similarity) and graph edit distance (PDG: semantic similarity). 
    
    \item CodeBLEU: A Method for Automatic Evaluation of Code Synthesis~\cite{ren2020codebleu}: uses a weighted combination of BLEU (4-gram), syntactic matching (matching subtrees in AST), and semantic matching (using dataflow structure). They demonstrate this has a better correlation with human evaluation compared to BLEU.
    \begin{align}
        \texttt{CodeBLEU} &= \alpha * \texttt{BLEU} + \beta * \texttt{BLEU}_{\texttt{weight}} +
        \gamma * \texttt{Match}_{\texttt{ast}} + \delta * \texttt{Match}_{\texttt{df}}
    \end{align}
    Where \texttt{BLEU} is the standard BLEU score~\cite{papineni2002bleu}, \texttt{BLEU}\_{\texttt{weight}} is the weighted n-gram match (high importance for keywords), \texttt{Match}\_{\texttt{ast}} is the syntactic AST match, and \texttt{Match}\_{\texttt{df}} is the semantic data flow match ~\cite{ren2020codebleu}
    Although codeBLEU captures multiple aspects of code compared to the BLEU score, it can measure the functional correctness of the code. Thus, codeBLEU alone is not a good measure for the evaluation of the generated source code.
    
    \item Exact Match (EM): Exact match compares the whole sequence to the ground truth. Given, programs without much overlap can produce the same results, EM is a harsh measure for the evaluation of source code. Exact match-based metrics are unable to account for large and complex space of program functions equivalent to the reference solution~\cite{chen2021evaluating}
    
    \item Edit distance: Levenschtein distance (minimum cost sequence of string edit operations to transform generated sequence to ground truth sequence) in case of source code sequence and graph edit distance~\cite{abu2015exact} (minimum cost sequence of node and edge edit operations to transform generated graph to ground truth graph) have been used in the literature for the evaluation of AST based code generation. As noted in the above metrics, it suffers similar problems and can not measure the functional correctness of the program.
    
    \item Unit tests: Unit tests are the set of conditions that need to be passed by the generated code. This metric is supported by test-driven development paradigm where developers first write unit tests before writing the software~\cite{li2022competition}. Given a sufficient number of unit tests that cover all the paths of execution of a reference code, we can verify that the two programs are functionally equivalent. But, limited test programs can falsely claim that the program is correct when it is not. Thus, sufficient unit tests are a really good method of evaluation of the generated source code. One problem with unit tests is that the generated code may be malicious. In such a case, it is better to run the generated code in the sandbox environment.

    \item pass@k metric~\cite{kulal2019spoc, chen2021evaluating}: This metric is used with unit tests, where the idea is to generate k samples per program and the problem is solved if any one of the k solutions passes the unit tests. But this one has high variance~\cite{chen2021evaluating}, and thus a modification was proposed by~\cite{chen2021evaluating} to generate $n > k$ solutions and count solutions that pass the unit tests as $c$. Then pass@k is defined as:
    \begin{align}
        pass@k &= \mathbb{E}_{\texttt{problems}} \Big[ 1 - \frac{{n-c \choose k}}{{n \choose k}} \Big]
    \end{align}
    
    also, timeout-based evaluations to approximate the algorithmic complexity of the generated solution~\cite{li2022competition} are being used in conjunction with the pass@k metric.
    
    \item CodeBERTScore and CodeScore: Evaluating Code Generation With Pretrained Models of Code~\cite{zhou2023codebertscore} and CodeScore: Evaluating Code Generation by Learning Code Execution~\cite{dong2023codescore} proposes to use LLM for generated code evaluation similar to BERTScore~\cite{zhang2019bertscore} that is used for evaluation of generated natural language. They claim that this metric has a higher correlation with human preference and with functional correctness than all existing metrics of code evaluation. It works by calculating the similarity of generated code and reference code in the embedding space where the embeddings are generated by using a pre-trained model called CodeBERT~\cite{feng2020codebert}. 
    
\end{itemize}

%% file: conclusion.tex
\section{Conclusion}
\label{section:conclusion}
Neural machine translation-based architectures are getting quite popular for source generation from various inputs. The NMT-based code generation is useful in multiple domains such as code generation from natural explanation, code generation from input binary or assembly (decompilation), code-to-code translation, code repair, bug fixing, and many more. In this survey paper, we cover the latest techniques being used for source code generation from multiple types of inputs. We also highlighted important techniques that are useful for source code generation along with the identification of current challenges and potential research directions. We presented common evaluation methods with their advantages and disadvantages. We think our survey papers lay the foundation for new researchers to start working in the field with an overall current set of techniques and motivate experienced researchers to develop new solutions that can solve the challenges identified in the paper and more.